\pdfoutput=1

\documentclass[11pt]{article}

\PassOptionsToPackage{table}{xcolor}
\usepackage[final]{acl}
\usepackage{comment}
\usepackage{times}
\usepackage{latexsym}
\usepackage{tabularx}
\usepackage{booktabs}
\usepackage{makecell}
\usepackage{placeins}
\usepackage{enumitem}
\newcommand*\samethanks[1][\value{footnote}]{\footnotemark[#1]}
\usepackage{subcaption}

\usepackage[T1]{fontenc}

\usepackage[utf8]{inputenc}

\usepackage{microtype}

\usepackage{inconsolata}

\usepackage{graphicx}

\title{Social Construction of Urban Space: 
\\ Using LLMs to Identify Neighborhood Boundaries From Craigslist Ads}

\author{\textbf{Adam Visokay \textsuperscript{$\dagger$}}\thanks{Equal contribution. Correspondence: \texttt{avisokay@uw.edu}} and
  \textbf{Ruth Bagley} \textbf{\textsuperscript{$\ddagger$}}\samethanks,
  \textbf{Ian Kennedy \textsuperscript{$\S$}},
  \textbf{Chris Hess \textsuperscript{$\P$}},
\\
  \textbf{Kyle Crowder \textsuperscript{$\dagger$}},
  \textbf{Rob Voigt \textsuperscript{$\ddagger$}},
  \textbf{Denis Peskoff \textsuperscript{$\ddagger$}}
\\
\\
  University of Washington, Department of Sociology \textsuperscript{$\dagger$}
\\
  Northwestern University, Department of Linguistics \textsuperscript{$\ddagger$}
\\
  University of Illinois Chicago, Department of Sociology \textsuperscript{$\S$}
\\
  Kennesaw State University, Department of Sociology and Criminal Justice \textsuperscript{$\P$}
\\}

\begin{document}
\maketitle

\begin{abstract}

Rental listings offer a window into how urban space is socially constructed through language.  
We analyze Chicago Craigslist rental advertisements from 2018 to 2024 to examine how listing agents characterize neighborhoods, identifying mismatches between institutional boundaries and neighborhood claims.
Through manual and large language model annotation, we classify unstructured listings from Craigslist according to their neighborhood.
Further geospatial analysis reveals three distinct patterns: properties with conflicting neighborhood designations due to competing spatial definitions, border properties with valid claims to adjacent neighborhoods, and ``reputation laundering" where listings claim association with distant, desirable neighborhoods.
Through topic modeling, we identify patterns that correlate with spatial positioning: listings further from neighborhood centers emphasize different amenities than centrally-located units.
Natural language processing techniques reveal how definitions of urban spaces are contested in ways that traditional methods overlook.

\end{abstract}

\section{Contested Neighborhood Boundaries}
\label{sec:intro}

Neighborhood location matters for a wide range of individual and collective outcomes \cite{sampson2002assessing, sharkey2014where, minh2017review, chyn2021neighborhoods}. Beyond objective demographic characteristics, the subjective features of a neighborhood—its reputation, status, or stigma—shape resident satisfaction, place attachment, and overall well-being \cite{tran2020my, kullberg2010does, permentier2011determinants, otero2024damages}. Neighborhood reputation also structures the economic value of property, patterns of investment, and the residential mobility that drives neighborhood stratification \cite{krysan2017cycle, evans2020neighborhood,kirk2024legitimising, korver2024good}.

\begin{figure}[t!]
    \centering
    \begin{subfigure}[t]{\linewidth}
          \raggedleft
        \includegraphics[width=.9\linewidth]{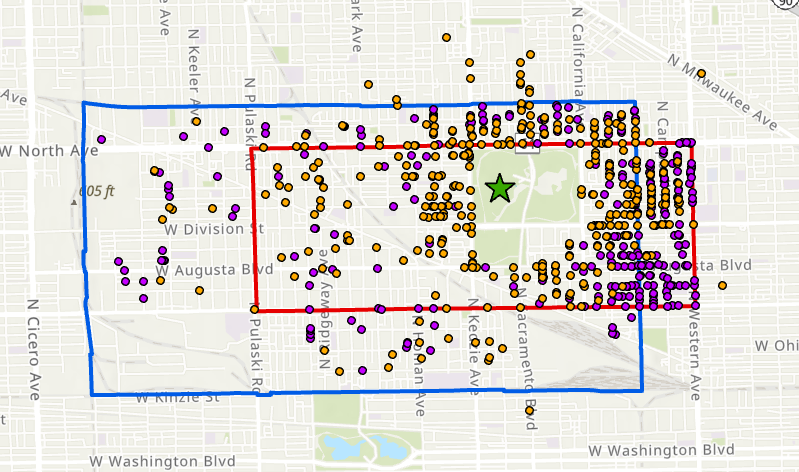}
    \end{subfigure}
    \begin{subfigure}[t]{\linewidth}
        \centering
        \includegraphics[width=\linewidth]{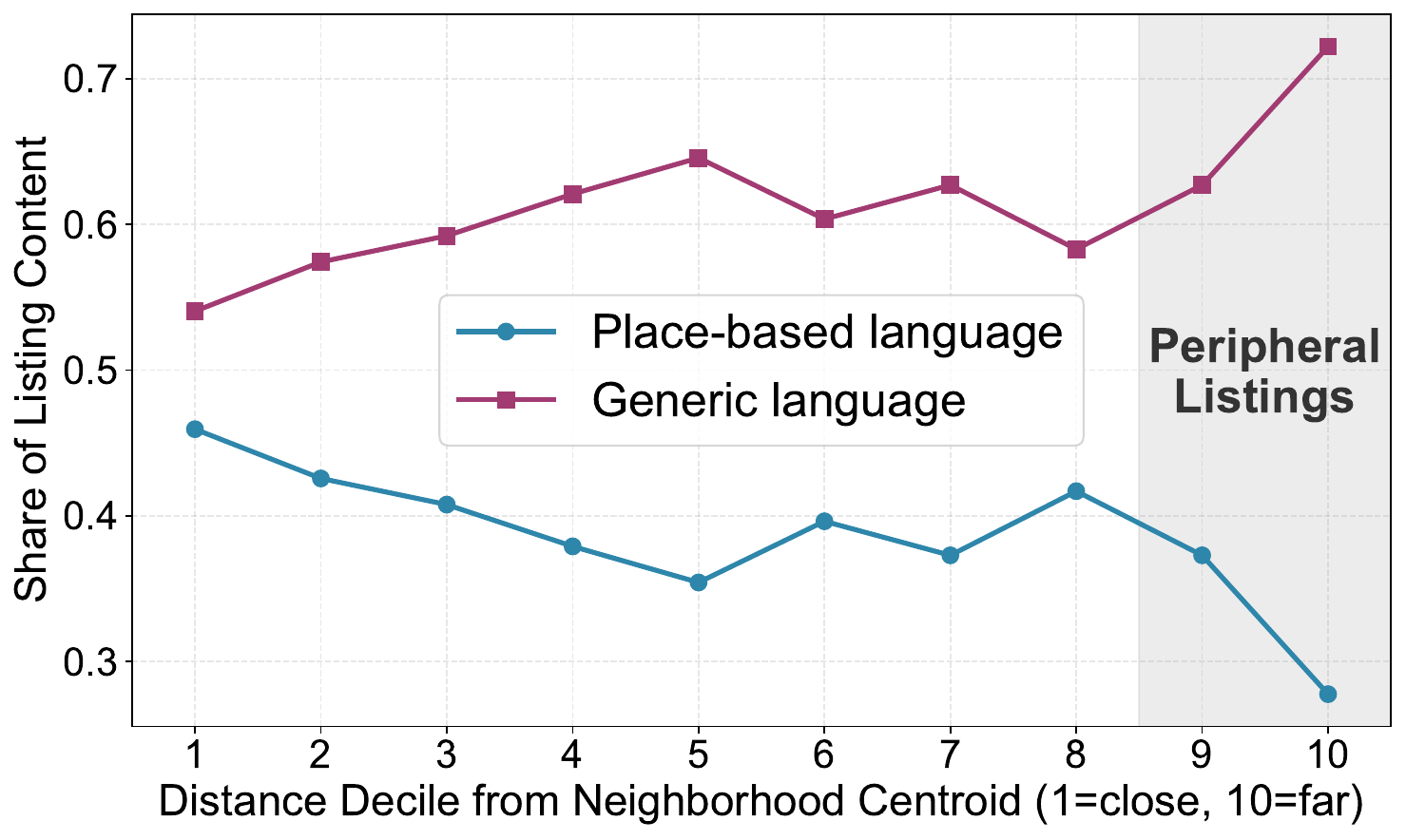}
    \end{subfigure}
    \caption{Extracting neighborhoods from unstructured rental listings with LLMs (RQ1, Section~\ref{sec:labeling}) provides insight into the social construction of space (\textit{RQ2}, top)\protect\footnotemark{} and allows us to study how language changes relative to distance from neighborhood centers (\textit{RQ3}, bottom).\protect\footnotemark{}}
    \label{fig:neighborhood_identity}
\end{figure}

      \footnotetext[1]{\textit{Conflicting conceptions} of the Humboldt Park neighborhood according to the official City of Chicago limits (blue), Zillow's boundary (red) and rental advertisements (points). Orange points depict unit listings claiming to be Humboldt Park while purple points claim elsewhere. The green star denotes the LLM-defined social center of Humboldt park.}
     \footnotetext[2]{Across Chicago, the share of place-based language decreases and generic ``boilerplate'' language increases in listings for units further from the social center of the neighborhood.}

We define \textbf{neighborhood identity} not just as a boundary in space, but as the spatial area that corresponds with the patterns of social activity and perceptions of people living in the area. This conceptualization builds on a lineage of research utilizing user-generated content (UGC) to define urban space ~\cite{plangprasopchok2009constructing, hollenstein2010,HiippalaLinguisticLandscape,BrunilaCriticalToponymy}. Because this identity is socially constructed through interaction and language, it is inherently fluid and contested. Identifying a singular ``true'' neighborhood boundary is not our aim, but rather, mapping the contours of contestation: the systematic slippage between institutional maps and the spatial claims made by social actors.

Neighborhoods are socially constructed through interaction and language \cite{zelner2015perpetuation, hohle2023rusty, stuart2024where}, yet traditional urban research often relies on rigid administrative boundaries as a proxy for neighborhood identity. In practice, listing agents frequently navigate a tradeoff between geographic fidelity and reputational leverage, substituting symbolic identity for physical proximity when properties sit at the periphery of desirable areas. We characterize this behavior not as random noise, but as a systematic distortion of urban space. Still, an important question remains: do spatial claims that depart from institutional maps always represent strategic ``reputation laundering" \cite{stuart2024where}, or might they reflect legitimate disagreements arising from the inherent ambiguity of socially constructed boundaries?

A related challenge for urban sociology has been the difficulty of observing and classifying such distortions at scale. Conventional data is blind to the strategic manipulation of spatial labels, and traditional NLP methods like string-matching struggle to distinguish between casual mentions and strong locational claims. Large Language Models (LLMs) provide a transformative opportunity to recover this latent social variable: claimed neighborhood identity. By evaluating LLMs on Chicago Craigslist rental advertisements (2018--2024), this work provides answers to the following Research Questions:
\vspace{5pt}
\begin{itemize}[nosep]
    \item \textbf{(RQ1) Measurement Viability:} Can zero-shot LLMs accurately identify specific neighborhood claims (vs. mere mentions) in unstructured rental listings compared to more traditional string-matching?
    \item \textbf{(RQ2) Social Location:} Where are neighborhoods actually located according to listing claims, and can we define a meaningful ``social center'' for each neighborhood?
    \item \textbf{(RQ3) Linguistic Substitution:} How does marketing language vary with spatial location? Specifically, does place-based language change as listings move farther from their claimed neighborhood center?
\end{itemize}\vspace{5pt}

Section~\ref{sec:data} describes our spatially-anchored corpus of Chicago listings. Section~\ref{sec:labeling} evaluates LLM performance against string-matching baselines \textbf{(RQ1)}. Section~\ref{sec:spatial} develops a framework for identifying neighborhood ``social centers'' \textbf{(RQ2)}. Section~\ref{sec:text} and Section~\ref{sec:regression} analyze the semantic structure and statistical associations between unit language and spatial positioning \textbf{(RQ3)}. Finally, Section~\ref{sec:results} contextualizes these findings within the broader field of computational social science.

\section{Craigslist Housing Advertisements}
\label{sec:data}

We use data collected from Chicago Craigslist rental advertisements from 2018 to 2024 to identify listings with mismatches between the neighborhood containing the unit and the neighborhood claimed by the listing agent.\footnote{We have been actively collecting data in Chicago since 2018, providing a rich window into the discursive construction of urban space. It includes all available advertisements each day from December 2018 until June 2024 using the webscraper \texttt{Helena} \cite{chasins2018rousillon, hess2022informing}. Occasional changes to the architecture of the Craiglist website result in limited periods of data loss, the longest of which was from August 2019 to early October 2019.}

\subsection{Why Craigslist?}

These rental listings offer a particularly valuable lens for examining how neighborhoods are socially constructed and contested because Craigslist's platform design creates an unusually unconstrained environment for spatial classification. Unlike many digital platforms that restrict users to predetermined administrative or commonly recognized neighborhood boundaries, Craigslist allows advertisers to freely designate any neighborhood label in their listings with unstructured text. This feature transforms rental advertisements into sites of boundary-making where the socio-spatial imaginary of the city becomes visible.

The neighborhood fields in these listings represent more than mere locational information—they reveal how real estate actors actively participate in constructing, reinforcing, or challenging existing spatial hierarchies. When landlords and property managers assign neighborhood labels to their listings, they engage in acts of spatial categorization that reflect both market strategies and internalized cognitive maps of urban space. These choices may align with officially recognized boundaries, reproducing understandings of place, or deliberately transgress established spatial categories to claim association with perceived higher-status areas.

Preceding computational analysis of Craigslist rental listings explores price distributions \cite{1CraigslistScraping}, neighborhood descriptions \cite{4SeattleRealEstate, 3RentalAvailability}, housing policy interventions \cite{boeing2021housing}, exclusionary language \cite{stewart2023move}, affordability \cite{hess2023segmented}, and home security \cite{somashekhar2024real}.   
Holistically, research shows that rental listings on Craigslist align with and appear to reproduce social inequality. 
Recent work has begun to focus on the importance of neighborhood names and the places those names describe \cite{schachter2024whats}, with specific focus on contested naming: when advertisements use neighborhood names that seem to diverge from the name most local residents would use for that space.

\subsection{Why Chicago?}

We focus on Chicago because it stands as a quintessential ``city of neighborhoods," where locally recognized community areas hold exceptional cultural, economic, and social significance \cite{hwang2014divergent}. Chicago is an ideal site for examining the social construction of urban space due to the high salience of its neighborhood boundaries. While the city maintains 77 officially recognized community areas, these rigid, non-overlapping boundaries often fail to represent the fluidity of neighborhood identity in reality. By using Chicago's stable institutional definitions as a point of comparison, we can more effectively identify and quantify how real estate actors use language to challenge or reinforce existing spatial hierarchies. This neighborhood orientation is so deeply embedded in Chicago's social fabric that it shapes how residents understand their place in the city, influences social networks, and structures daily mobility patterns~\cite{nyt2024rentrenewal}.

By analyzing patterns in how these spatial designations align with or diverge from official boundaries, we can observe in real time the processes through which neighborhood reputations are reinforced or contested. The negotiated quality of these spatial boundaries becomes particularly visible when examining cases where advertisers claim association with neighborhoods other than those in which units are formally located, according to administrative boundaries. Such instances of spatial repositioning offer a window into the dynamics that shape how urban space is valued, categorized, and ultimately experienced by various stakeholders from listing agents to residents or potential tenants to municipal administrators.

\section{Detecting Neighborhoods  with LLMs}
\label{sec:labeling}

Online rental advertisements are generally unstructured and vary widely between listings. Distinguishing between which neighborhoods are mentioned in a listing from which neighborhood(s) the listing claims to be in is a nuanced task of great importance to social scientists interested in the social construction of urban space. This answer to ``which neighborhood does this advertisement claim the unit to be in?" is not always obvious, even to a human annotator. 
For example, 
\begin{quote}
    \dots this fully restored \textbf{East Lakeview} property sits on a beautiful tree-lined street located in the heart of the popular \textbf{Wrigleyville} neighborhood \dots
\end{quote}

\noindent It is clear based on the language that this advertisement is not merely mentioning these neighborhoods, but staking a strong claim to being located in both. Wrigleyville is a Chicago neighborhood within the larger neighborhood of East Lakeview, so this claim is entirely coherent. However, reconciling such competing claims at scale is a principal challenge inherent to this particular task.

Furthermore, some listings contain mentions of several irrelevant neighborhoods, even dozens like this example:

\begin{quote}
\dots
Disclaimer: Pricing, availability, and specials are subject to change at any time and without notice.
HotSpot Rentals services the following neighborhoods: South Loop, Printers Row, Near North, River North, Gold Coast, West Loop, Fulton River District, West Loop Gate, The Loop, Streeterville, Lakeshore East, New East Side, Old Town, Medical District, University Village, Near North, River West, Lincoln Park South, Lakeview, Uptown, Ukrainian Village, Wicker Park, Edgewater, Ravenswood, Bronzeville, Logan Square.
\end{quote}

Making the distinction between neighborhood mentions and strong claims that a unit is in a particular neighborhood is a nuance that large language models are particularly well-suited for compared to existing methods. To make this comparison, first we label the full corpus using a bespoke string-matching approach which serves as the baseline ``best practice" which we compare to the Language Model-based labeling. Both sets of labels are evaluated against a subset of 200 manually labeled neighborhood listings. These manual labels were produced by authors of this article. 

\subsection{Manual Labeling}

The process for creating our validation set of ``gold standard" labels considers three sources of information for each advertisement in the following order. First, if the title field includes a neighborhood name, that becomes the manual label for the neighborhood claim. Then if there is no claim in the title, we consider the body of the listing. This is the largest source of text in each advertisement, and also the most ambiguous with respect to identifying strong claims. When faced with multiple claims -- as in the quote above -- we take the first claim as the manual label. Finally, if neither the title nor body fields contain a neighborhood claim, we extract a neighborhood claim from the neighborhood field, if it exists. An advertisement only receives the `unknown' label if there is not a strong claim in any of the three fields. We follow the same prioritization scheme in the string-matching and LM labeling procedures.

\subsection{Data Pre-Processing}

Before labeling we performed standard data pre-processing and de-duplication on the raw text. We removed common boilerplate text that appeared in virtually all Craigslist listings (such as ``QR Code Link to This Post''), corrected Unicode translation errors to ensure consistent character rendering, and precisely mapped listings onto Zillow and City of Chicago neighborhood boundaries to confirm geographical validity. We also performed thorough de-duplication of listing entries by title and body text, retaining only the most recent version when multiple entries existed, as Craigslist saves edited listings as separate posts. This preparation distilled a clean dataset of 30,531 unique listings from an initial corpus of n=128,764 initial observations.

\subsection{String-Match Labeling Neighborhoods }

To determine which Chicago neighborhood a rental advertisement belongs to based on the text in the post we begin with a list of 192 distinct neighborhoods from Zillow, a real-estate marketplace company which provides commercial neighborhood lists for major US cities. Then, we manually construct a comprehensive list of regular expressions that can match the official name and its alternatives (e.g. \textit{wriggleyville, wriglyville, wrigglyville, wrigley ville} for \textit{wrigleyville}). These regex patterns are designed to be flexible with spaces and case-insensitive. Following the same protocol as the manual annotations, we use a function which tries to match neighborhoods in the listing title, body and dedicated neighborhood field. When multiple neighborhoods match, we select the label which appears earliest in the text. 

\subsection{Language Model Labeling Neighborhoods}

We prompt GPT-4.1 mini as a high quality relative to cost option.\footnote{We compare a sample of regular vs mini vs nano, and other non-OpenAI options.  Reasoning models are unnecessary for this extraction task.} 
Table~\ref{tab:prompt} contains our prompt.

\begin{table}[ht]
\small
\rowcolors{2}{gray!10}{white}
\setlength{\tabcolsep}{6pt}
\begin{tabular}{>{\ttfamily}p{.9\columnwidth}}
\rowcolor{gray!30}
\textbf{BASE\_PROMPT} \\
\hline
Extract the Chicago neighborhood from the rental text. \\

Rules: \\
- NEVER use the address or zip code to determine the neighborhood \\
- Choose only explicitly stated neighborhood from possible responses \\
- If neighborhood is unclear mark it as [unknown] \\
- Format precision: ``lakeview" but not ``x, y, z" for ``lakeview residence near x, y, z" \\

text: \{body\} \\
Possible responses: \{zillow\_list\} \\
Return only: \\
label: [neighborhood] \\
\hline
\end{tabular}
\caption{Prompt format for extracting Chicago neighborhood claims from rental listings.}
\label{tab:prompt}
\end{table}

We create three separate labeling workflows, one for each title, body and neighborhood field. 
We input the data independently to avoid leakage.
A Zillow neighborhood list is provided to constrain possible responses to items in the list.
In the case of unknowns, we prioritize the label from the title, then the body, then the neighborhood field.
Only if all three are unknown is the final neighborhood claim labeled `unknown'. 

\subsection{Label Post-Processing}

While we engineer the prompt to provide structured output in the form \texttt{label:response}, the outputs still require post-processing. 
We implement a multi-stage post-processing pipeline to standardize the three LLM labels for each advertisement and assess model performance against manual validation.
Through manual review we flag instances of minor spelling discrepancies (e.g. `lakeview` instead of `lake view`, `wrigglyville` instead of `wrigleyville`). 
For text normalization, we construct a replacement dictionary to correct such instances. We implement the same priority-based claim-selection algorithm as is used for manual annotation and string matching. 
The post-processing system first examines the listing title label; if no neighborhood is identified, it analyzes the listing body label, and if still unsuccessful, it checks the neighborhood field label; and only then returns 'unknown' classification. 
This aligns with how potential renters process listing information, prioritizing the most prominent textual elements. 
When faced with compound neighborhood designations, we prioritize the first neighborhood when multiple were present, just as we do for manual labeling and string-matching. For outputs that contain separators (e.g. `uptown, ravenswood` or `avondale/logan square`), we extract the first neighborhood claim before a separator. 

\subsection{Evaluation of Labeling Task}

We compare the string-match and LLM labels to the manual labels in the 200-item validation set. Performance metrics are shown in Table \ref{tab:label_metrics}.
While accuracy scores are comparable (GPT-4 mini's 85\% is marginally better than the string matching's 79\%), the disagreements largely reflect the inherent ambiguity of neighborhood classification -- a challenge even human annotators face when listings claim multiple neighborhoods. 
A notable advantage of the LLM approach is its efficiency and scalability.
Unlike string matching on keywords, which requires extensive manual configuration of locale specific patterns, the pre-trained LLM can work with a zero-shot prompt, making it adaptable. 

\begin{table}[t]
\small
\centering
\begin{tabularx}{\columnwidth}{@{}lcc@{}}
\toprule
\textbf{Metric} & \textbf{String Match} & \textbf{GPT-4.1} \\
\midrule
Accuracy & 0.79 & \textbf{0.85} \\
Macro Average F1 & 0.62 & \textbf{0.70} \\
Weighted Average F1 & 0.77 & \textbf{0.85} \\
\bottomrule
\end{tabularx}
\caption{GPT-4.1 outperforms the string match-based keyword search on the neighborhood claim labeling task. While the performance gain is marginal, the LLM labeling process was cheap, fast and can be scaled up.}
\label{tab:label_metrics}
\end{table}

\section{Geospatial Analysis of Neighborhoods}
\label{sec:spatial}

\begin{figure}[t!]
    \centering
    \includegraphics[width=.9\linewidth]{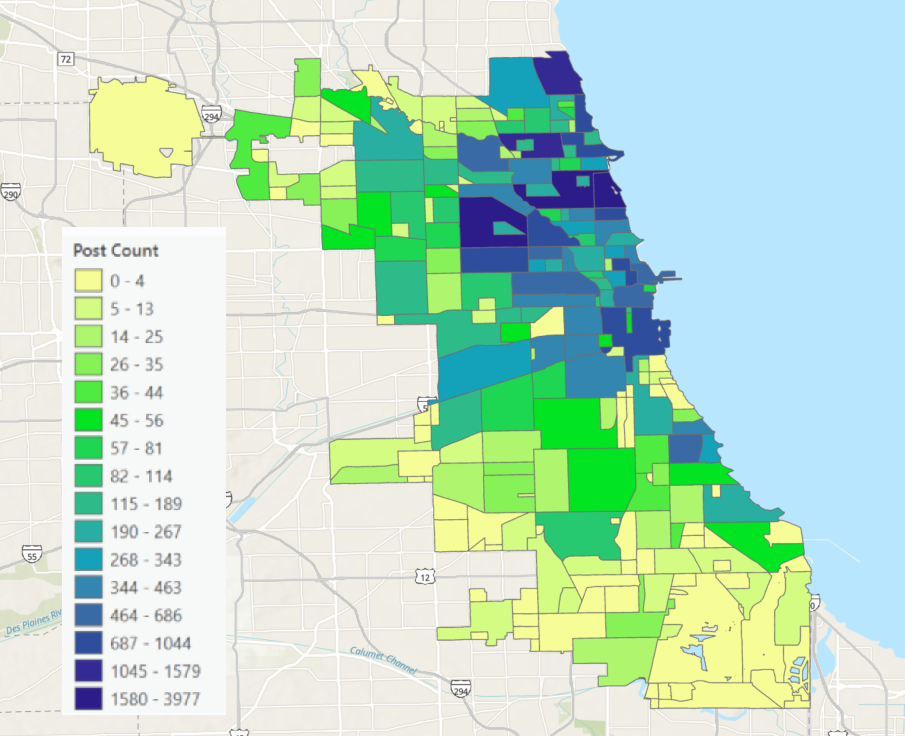}
    \caption{The city of Chicago and its constituent neighborhoods, as defined by Zillow, colored to represent the number of posts for properties located in each area}
    \label{fig:heatmap}
\end{figure}

Although Craiglists rental listings do not comprehensively show every available property in the Chicago area, there is still substantial coverage across the city, as seen in Figure \ref{fig:heatmap}. As a result, we take these rental listings to be a reasonably representative sampling, which we use to identify neighborhoods as they might be conceived of by the people, or at least the realtors, of Chicago.

Neighborhood boundaries are nebulous and hard to define; even the City of Chicago and Zillow, both with access to a great deal of data/information, have developed quite different maps of neighborhood boundaries. 
Figure \ref{fig:neighborhood_identity} shows that while the official Zillow and city boundaries of Humboldt Park include most of the posts claiming to be in that neighborhood, none of the borders are the same. In addition, borders between neighborhoods are likely less rigid than an official boundary might suggest; despite being located within the formal boundaries of Humboldt Park, there are a number of listings, mainly around the edges, claiming to be part of other nearby neighborhoods, primarily Ukrainian Village, Wicker Park, or West Town.

Using the rental listings, we conceptualize the "social center" of the neighborhood as the centroid of all listings that claim to be located within that neighborhood. We use \texttt{geopandas} to identify the centroid of all posts claiming to be in the same neighborhood using the latitude and longitude coordinates of all property locations. The map of Humboldt Park suggests this is an effective method, because the social center (represented by the green star) is in fact located in the eponymous park which is considered to be the heart of the neighborhood.

However, not all posts claiming to be in a given neighborhood may have the same strength to their claim; advertisements for apartments at the center of a popular neighborhood like Logan Square likely have different characteristics than posts for units on the fringes of the neighborhood. The posts on the fringes might even be trying to seem more desirable by claiming to be in a more popular neighborhood, whereas it might be more of an objective description of location for a property actually located on Logan Square. We conceptualize this by identifying how far from the social center a post is, using three metrics for distance: 1) Raw Distance: Euclidean distance between the latitude and longitude coordinates of the post and that of the neighborhood centroid; 2) Relative Distance:  raw distance for all posts in the neighborhood projected onto a [0,1] interval using min-max scaling; 3) Z-scored Distance:  z-scored distance for posts claiming to be in the same neighborhood

We use these measures to distinguish a specific subset of the posts: those on the periphery of a neighborhood. We define peripheral posts as those that are in the furthest 20\% of posts from the centroid for a given neighborhood (for any neighborhood labels with at least 5 posts).

\subsection{Comparison of Neighborhood Boundaries}

In order to get a more concrete understanding of different conceptions of neighborhoods, we identify two more neighborhoods to explore more in depth: Logan Square and Pilsen. 

Logan Square is a well-known neighborhood in Chicago, and also a neighborhood label where a large number (n=2495) of listings claim to be. We can see in Figure \ref{fig:logan} that the City of Chicago boundaries for Logan Square contain primarily postings claiming to be in the neighborhood (orange points) without many posts claiming to be elsewhere (purple). The Zillow boundaries do encompass Logan Square claims that the Chicago boundaries do not, but the areas excluded by Chicago also have a higher concentration of claims of being in other neighborhoods. In addition, there are a number of listings claiming to be in Logan Square that are outside both formal boundaries--these posts would certainly be part of the 'peripheral' posts we defined earlier. This kind of posting merits further exploration to better understand what is happening in listings that claim to be part of a neighborhood when that might be more likely to be contested.

\begin{figure}[t!]
    \centering
    \includegraphics[width=\linewidth]{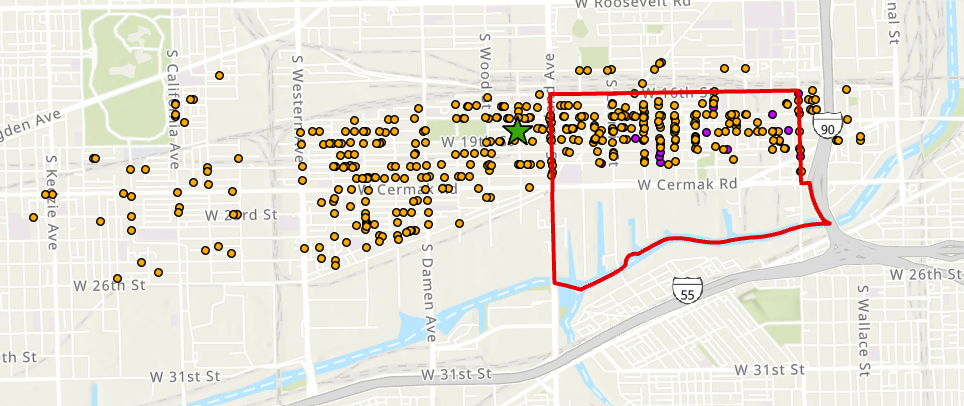}
    \caption{Contested boundaries for Pilsen neighborhood according to Zillow's definition (red). Orange points depict listings claiming to be in Pilsen, purple points are listings claiming to be elsewhere). The green star depicts the Pilsen neighborhood centroid. This is an example of what we call \textit{border stretching}, demonstrating how static boundary systems may not capture the neighborhood identities as experienced by the people living in them.}
    \label{fig:pilsen}
\end{figure}

Although the blue City of Chicago boundaries appear to be a better approximation for Logan Square in Figure \ref{fig:logan}, this is not the case for every neighborhood. Pilsen, shown in Figure \ref{fig:pilsen}, is another well-known neighborhood within Chicago, but it is not included as a distinct neighborhood by the City of Chicago. However, even though Zillow does include Pilsen as an area, it also does not seem to define it in the same way as Craigslist advertisements. Many of the posts claiming to be in Pilsen appear in a clustered way outside of the Zillow boundaries, and even the centroid of the Craigslist-defined neighborhood is outside the Zillow bounds.
This could represent a developing neighborhood identity, which may not have been as strong at the time of the map creation, and supports the need for a more dynamic model of neighborhoods.

\section{Content Analysis of Rental Listing Text}
\label{sec:text}

We use the \texttt{tomotopy} Python package to identify latent topics in the content posted in Craigslist rental advertisements. \texttt{tomotopy} uses Gibbs-sampling and is based on the LDA approach described in \cite{blei2003latent} and \cite{newman2009distributed}. We prepared the text corpus by combining listing titles and body text from the Craigslist dataset. Standard preprocessing was applied using NLTK -- lowercasing, removing alphanumeric characters, tokenization, custom stopword filtering, and lemmatization. We remove common real estate jargon that would otherwise dominate the topic distributions without providing meaningful differentiation between the content in different listing types. For the LDA implementation we selected hyperparameters that allow for moderate document sparsity ($\alpha = 0.1$) and greater topic-word concentration ($\eta = 0.01$). We trained for 100 iterations and tried $k=5, 7$ and $10$ topics which returned coherence scores of \texttt{0.7344}, \texttt{0.7425} and \texttt{0.7509}, respectively. After manual review we decided to focus on the $k=7$ results for our remaining analysis as these had clearer separation than the $k=5$ topics and are more interpretable than the $k=10$ results. 

\subsection{Topic Interpretations}

We identify seven distinct topics in the Craigslist advertisements, shown in Table \ref{tab:topic_words}. These topics reveal distinct patterns in how rental listings are framed, which have important implications for understanding the conception of neighborhood reputation and representation in the online rental market. 

\begin{table}[t]
\small
\centering
\begin{tabularx}{\columnwidth}{@{}lX@{}}
\toprule
\multicolumn{1}{c}{\textbf{Topics}} & \multicolumn{1}{c}{\textbf{Common Words}} \\
\midrule
\makecell[tl]{\textbf{1.} Furnished Short-Term \\ \hspace{9pt}\textit{3.6\% of corpus}} & lease, home, month, furnished, mo, amenity, community, view, apartment, offer, blueground, access, neighborhood, enjoy, stay, loop \\
\makecell[tl]{\textbf{2.} Rental Terms \\ \hspace{9pt}\textit{25.9\% of corpus}} & fee, lease, included, tenant, credit, street, pay, deposit, application, large, move, heat, per, gas, security, pet, utility \\
\makecell[tl]{\textbf{3.} Property Search \\ \hspace{9pt}\textit{1.1\% of corpus}} & apartment, rental, property, place, hill, cheap, grove, find, apt, height, center, agency, search, local, credit \\
\makecell[tl]{\textbf{4.} Spanish Language \\ \hspace{9pt}\textit{2.1\% of corpus}} & apartment, rental, property, place, hill, cheap, alquilar, propiedades, buscar, alquileres, height, search, agency, google \\
\makecell[tl]{\textbf{5.} Amenities \\ \hspace{9pt}\textit{35.0\% of corpus}} & floor, central, new, appliance, dishwasher, heat, large, space, air, stainless, feature, modern, steel, living, cat, closet \\
\makecell[tl]{\textbf{6.} Listing Conditions \\ \hspace{9pt}\textit{13.3\% of corpus}} & subject, unit, change, price, center, property, onsite, hour, availability, special, dog, pricing, studio, fitness, housing, amenity \\
\makecell[tl]{\textbf{7.} Contact Information \\ \hspace{9pt}\textit{18.9\% of corpus}} & contact, info, show, click, feature, view, call, id, renovated, , included, closet \\
\bottomrule
\end{tabularx}
\caption{Topic interpretations based LDA topic modeling on Chicago Craigslist rental listings.}
\label{tab:topic_words}
\end{table}

Topic 1 focuses on furnished short-term rentals, highlighting amenities and comfort with words like ``furnished," ``month," and ``stay," suggesting a market segment catering to temporary residents seeking turnkey living situations. 
Topic 2 centers on rental requirements and financial considerations, with terms like ``fee," ``credit," ``deposit," and ``application," reflecting the administrative and financial aspects of renting. 
Topics 3 and 4 both relate to property search, with Topic 4 specifically including Spanish-language terms like ``alquilar" and ``propiedades," indicating efforts to reach Spanish-speaking audiences in Chicago's rental market. Topic 5, the most prevalent across the corpus at approximately 35\% of document content, focuses on apartment features and amenities such as ``appliance," ``stainless," ``modern," and ``dishwasher," underscoring the prevalence of interior quality in marketing rental properties. 
Topic 6 addresses listing conditions and availability, featuring terms related to pricing, special offers, and property policies. Finally, Topic 7 concentrates on contact information and viewing arrangements, with words like ``contact," ``show," ``click," and ``schedule," facilitating the connection between potential renters and property managers. 
The distribution of these topics across advertisements reveals how Chicago's rental market is presented online, with physical features and financial terms dominating the discourse.
\subsection{Regression Analysis}
\label{sec:regression}

We estimate the relationship between listing characteristics and proximity to the social center of an associated neighborhood using OLS regression with relative distance to neighborhood center as our primary dependent variable. Our model includes unit characteristics (bedrooms, bathrooms, rent, and square footage) and the topic proportions identified in our LDA analysis. For a full table of regression outputs see Table \ref{tab:regression_results} in the Appendix.

This analysis reveals several patterns in spatial representations. Advertisements with more bedrooms/bathrooms are associated with being further from neighborhood centers, while higher-priced listings are closer to their respective social centers. Most notably, listings with higher proportions of Topic 3 (Property Search) content exhibit increased relative distance from the center of the neighborhood (+0.20, p$<$0.001).
Conversely, listings emphasizing apartment amenities (Topic 5) are associated with being closer to the centroid (-0.03, p$<$0.01). These findings indicate that misrepresentation is not random but correlates with specific marketing approaches in rental listings.

\section{Results}
\label{sec:results}

Our analysis of over 30,000 unique Chicago rental listings reveals that neighborhood boundaries are actively renegotiated through strategic linguistic claims. By establishing a ``social center'' for neighborhoods based on the density of textual claims rather than rigid municipal boundaries, we demonstrate how agents navigate the tradeoff between geographic reality and reputational leverage. The following sections detail our findings in relation to our primary research questions.

\subsection{Labeling Viability (RQ1)}
We find that Large Language Models are highly effective for identifying neighborhood claims within unstructured text, outperforming traditional string-matching techniques. While the accuracy gain is incremental (85\% for GPT-4.1 mini vs. 79\% for string-matching), the LLM approach can scale beyond Chicago to other urban contexts without requiring reconfiguration or the same level of local real estate knowledge. Our labeling system prioritizes precision by selecting the single strongest neighborhood claim, addressing the inherent ambiguity found in listings that mention multiple areas.

\subsection{Defining Social Centers (RQ2)}
Our geospatial analysis reveals that the ``social center'' of a neighborhood—defined as the centroid of all property listings claiming that identity—often aligns with local landmarks, such as the eponymous park in Humboldt Park. However, these Craigslist-defined centers frequently diverge from institutional boundaries. We identify significant variation in representation: neighborhoods like Lake View are heavily overrepresented in claims relative to their Zillow-defined footprints, while areas such as Englewood and North Austin are significantly underrepresented, suggesting a lack of strong neighborhood identity or the presence of territorial stigma in the rental market.

\subsection{Linguistic Substitution (RQ3)}

Our analysis characterizes spatial misrepresentation not as random market noise, but as a predictable distortion of urban space. We identify three distinct patterns of spatial claim discrepancies: (1) \textit{Conflicting Conceptions}, where stakeholders disagree on boundaries (e.g. Humboldt Park in Figure~\ref{fig:neighborhood_identity}); (2) \textit{Border Stretching}, where listings claim adjacent, plausible neighborhoods (e.g. Pilsen in Figure~\ref{fig:pilsen}); and (3) \textit{Reputation Laundering}, where properties or peripheral areas claim association with distant, desirable neighborhoods.

To identify these patterns systematically, we operationalize ``peripheral claims" as those located in the furthest 20\% from a neighborhood’s social centroid. By comparing these LLM-labeled claims against institutional Zillow boundaries, we find evidence of systematic over and under-representation. Specifically, Lake View is significantly overrepresented--claimed more frequently than geographic distributions would predict--while neighborhoods such as Englewood, North Austin, and Hanson Park are significantly underrepresented. This pattern is consistent with reputation laundering: agents appear to distance properties from stigmatized neighborhood names while strategically claiming higher-status alternatives.

This substitution is statistically observable through a compensatory language pattern. Regression results show that as properties move further from the neighborhood social center, listing agents substitute symbolic identity for physical proximity. For instance, generic property search language (Topic 3) is positively associated with relative distance from centroid ($+0.20, p < 0.001$), while specific interior amenity language (Topic 5) is negatively associated ($-0.03, p < 0.01$). Further, compared to central listings, the language used in peripheral listings shifts from location specific, non-portable amenities (Topics 1,5) toward more abstract and generic property-search language (Topics 2,3,4,6,7), a pattern illustrated in Figure \ref{fig:neighborhood_identity}. 
\section{Implications Beyond Sociology}
\label{sec:discussion}

The use of LLMs has exploded in recent years, and they can be seen by the general public as a simple, reliable solution to many routine problems. However, it remains an open question how powerful they may be in interdisciplinary research~\citep{ziems2024can}. In order to better understand the impact of NLP on a broader scale and help address a specific question in the field of Urban Sociology, we demonstrate the effectiveness of LLMs at a notoriously difficult task: identifying where rental listings claim to be located on a large scale, in order to inform our understanding of processes of social construction of urban space.

However, although the impact of language models on this task may be transformative in the ability to quickly expand the scope of analysis, the quantifiable improvement on simple algorithms designed by experts is perhaps more incremental. In addition, the LLM labeling was certainly not perfectly accurate, suggesting that there is still room for improvement in large language models that might not be captured by tests and benchmarks developed solely within the field of NLP, and that interdisciplinary collaboration could lead to improvements both in NLP methodology and in making research questions and analyses in a broad range of fields more tractable and scalable.

\newpage

\section{Limitations}

Our analysis of Craigslist rental listings provides valuable insights into neighborhood claims and social construction, but several important limitations should be acknowledged. The process of collecting data from Craigslist presents inherent challenges regarding post uniqueness and identification. Despite our deduplication efforts, the platform's structure makes it difficult to definitively determine which posts represent truly unique listings versus slightly modified versions of the same property.

While our dataset offers substantial coverage across Chicago neighborhoods, it cannot claim to be fully representative of the entire rental market. Craigslist represents just one segment of available rental properties, potentially skewing toward certain price points or property types. Additionally, our data may over represent certain management companies and realtors who post frequently on the platform, as opposed to ``mom and pop" owners. These high-volume posters are more likely to use standardized language across multiple listings, which may introduce uniformity in how neighborhoods are described that doesn't reflect broader market patterns.

A fundamental challenge in this research is the absence of an authoritative catalog of Chicago's neighborhoods. As we argue, such a catalog is conceptually impossible. For practical purposes, we relied on Zillow's neighborhood boundaries as our primary reference, but these designations do conflict with local understandings. For example, questions arise about whether "West Loop" constitutes its own neighborhood distinct from "West Loop Gate," or whether "East Rogers Park" should be considered separate from "Rogers Park." These ambiguities reflect the socially constructed nature of neighborhoods themselves. Also, many units along Lake Michigan have a view of the water, and therefore advertise ``views of the lake" or ``lake views" which can be impossible to distinguish from listings claiming to be a ``lake view" without considering the corresponding latitude and longitude coordinates. 

Furthermore, assigning a single neighborhood label to each listing proved challenging when advertisements contained multiple, sometimes competing neighborhood claims. Our hierarchical labeling approach (prioritizing title, then body, then neighborhood field) necessarily simplifies what can be complex spatial positioning strategies employed by listing agents. A rental advertisement might strategically claim association with multiple neighborhoods simultaneously, a nuance our single-label framework cannot fully capture. For instance, a listing located on the border between Logan Square and Humboldt Park might leverage both neighborhood identities depending on the perceived audience and market conditions.

These limitations highlight the inherent complexity of studying socially constructed spatial boundaries through digital traces and suggest opportunities for future research employing more nuanced approaches to neighborhood identification and classification.

\section*{Ethics Statement}
Deciding how to approach this analysis is a non-trivial decision and following the extensive work in sociology and economics is important.  We spoke experts in both fields in scoping and executing this project. 

We collect data from Craigslist which, in some cases, contains specific information about individual posters. Craigslist is a public forum whose housing section should not contain much information irrelevant to the housing ads themselves. We do not release these data publicly per the non-commercial use terms of service and ensure no PII appear in any of our writing, results or figures. 

Another limitation when working with pretrained foundation models like GPT-4 is a lack of reproducibility, as we do not have access to the training data or the weights. In the interest of reproducibility, we keep annotation costs under \$100. 

We utilized multiple Generative AI tools (OpenAI’s GPT-4/5 and Anthropic’s Claude 4.5 Opus/Claude Code) in the production of this manuscript, in the following ways: 1) producing computer code for data cleaning and analysis and 2) iteratively improving the concision and clarity of the writing. We have carefully reviewed all aspects of the manuscript for accuracy and coherence. All scientific insights, analysis and interpretation of data and scientific conclusions are made solely by the authors. 

\section{Acknowledgements}

We thank Drew Messamore, Diag Davenport, and Michael Reher for the providing valuable suggestions on an earlier version of this manuscript. Adam gratefully acknowledges the resources provided by the International Max Planck Research School for Population, Health and Data Science (IMPRS-PHDS). We are also grateful to the University of Washington’s Department of Sociology writing workshop for their comments and feedback. 

\clearpage



\clearpage
\appendix
\onecolumn

\section{Full Performance Metrics for Neighborhood Claim Validation}
\label{sec:appendix}
\begin{table*}[b!]
\small
\centering
\setlength{\tabcolsep}{4pt}
\begin{tabular}{@{}lccccccr@{}}
\toprule
& \multicolumn{3}{c}{\textbf{String Match}} & \multicolumn{3}{c}{\textbf{GPT-4 Mini}} & \\
\cmidrule(lr){2-4} \cmidrule(lr){5-7}
\textbf{Neighborhood} & \textbf{Prec.} & \textbf{Recall} & \textbf{F1} & \textbf{Prec.} & \textbf{Recall} & \textbf{F1} & \textbf{Support} \\
\midrule
the loop & 0.12 & 0.50 & 0.20 & 1.00 & 1.00 & 1.00 & 2 \\
rogers park & 1.00 & 1.00 & 1.00 & 0.75 & 1.00 & 0.86 & 3 \\
lake view & 0.63 & 0.86 & 0.73 & 0.94 & 0.73 & 0.82 & 22 \\
ranch triangle & -- & 0.00 & 0.00 & -- & 0.00 & 0.00 & 1 \\
lincoln park & 0.90 & 1.00 & 0.95 & 0.94 & 0.89 & 0.91 & 18 \\
fulton river district & -- & 0.00 & 0.00 & 1.00 & 1.00 & 1.00 & 1 \\
south loop & 1.00 & 1.00 & 1.00 & 1.00 & 1.00 & 1.00 & 3 \\
park west & -- & 0.00 & 0.00 & -- & 0.00 & 0.00 & 1 \\
west town & 1.00 & 1.00 & 1.00 & 1.00 & 1.00 & 1.00 & 1 \\
logan square & 0.88 & 0.88 & 0.88 & 0.89 & 1.00 & 0.94 & 8 \\
lake view east & -- & 0.00 & 0.00 & 0.90 & 0.90 & 0.90 & 10 \\
university village - little italy & -- & 0.00 & 0.00 & -- & 0.00 & 0.00 & 2 \\
uptown & 1.00 & 0.83 & 0.91 & 1.00 & 0.83 & 0.91 & 6 \\
wicker park & 1.00 & 1.00 & 1.00 & 1.00 & 1.00 & 1.00 & 4 \\
portage park & 1.00 & 1.00 & 1.00 & 1.00 & 1.00 & 1.00 & 2 \\
old town & 1.00 & 0.83 & 0.91 & 1.00 & 0.83 & 0.91 & 6 \\
avondale & -- & 0.00 & 0.00 & -- & 0.00 & 0.00 & 1 \\
west loop gate & -- & 0.00 & 0.00 & 1.00 & 1.00 & 1.00 & 4 \\
streeterville & 1.00 & 1.00 & 1.00 & 1.00 & 1.00 & 1.00 & 2 \\
ravenswood & 1.00 & 0.80 & 0.89 & 1.00 & 0.80 & 0.89 & 5 \\
wrigleyville & 1.00 & 0.67 & 0.80 & 1.00 & 1.00 & 1.00 & 3 \\
river north & 0.83 & 0.71 & 0.77 & 0.86 & 0.86 & 0.86 & 7 \\
buena park & 1.00 & 1.00 & 1.00 & 0.75 & 1.00 & 0.86 & 3 \\
bucktown & 1.00 & 1.00 & 1.00 & 1.00 & 1.00 & 1.00 & 7 \\
kenwood & -- & 0.00 & 0.00 & 1.00 & 1.00 & 1.00 & 1 \\
old irving park & 1.00 & 1.00 & 1.00 & 1.00 & 1.00 & 1.00 & 1 \\
edgewater & 1.00 & 1.00 & 1.00 & 1.00 & 1.00 & 1.00 & 5 \\
pilsen & 1.00 & 1.00 & 1.00 & 1.00 & 1.00 & 1.00 & 7 \\
garfield ridge & 1.00 & 1.00 & 1.00 & 1.00 & 1.00 & 1.00 & 1 \\
humboldt park & 1.00 & 1.00 & 1.00 & 1.00 & 1.00 & 1.00 & 1 \\
andersonville & 1.00 & 1.00 & 1.00 & 1.00 & 0.67 & 0.80 & 3 \\
west rogers park & 1.00 & 1.00 & 1.00 & 1.00 & 1.00 & 1.00 & 1 \\
ukrainian village & 1.00 & 1.00 & 1.00 & 1.00 & 1.00 & 1.00 & 1 \\
gold coast & 0.75 & 1.00 & 0.86 & 0.75 & 1.00 & 0.86 & 3 \\
hyde park & 1.00 & 1.00 & 1.00 & 1.00 & 1.00 & 1.00 & 1 \\
roscoe village & 1.00 & 1.00 & 1.00 & 0.33 & 1.00 & 0.50 & 2 \\
east garfield park & 1.00 & 1.00 & 1.00 & 1.00 & 1.00 & 1.00 & 1 \\
albany park & 1.00 & 1.00 & 1.00 & 1.00 & 1.00 & 1.00 & 1 \\
lincoln square & 0.33 & 1.00 & 0.50 & 0.00 & 0.00 & 0.00 & 1 \\
south shore & 1.00 & 1.00 & 1.00 & 1.00 & 1.00 & 1.00 & 1 \\
hermosa & 1.00 & 1.00 & 1.00 & 0.50 & 1.00 & 0.67 & 1 \\
ravenswood manor & -- & 0.00 & 0.00 & -- & 0.00 & 0.00 & 1 \\
unknown & 0.78 & 0.88 & 0.82 & 0.50 & 0.50 & 0.50 & 8 \\
\midrule
\textbf{Average Metrics} & & & & & & & \textbf{Total: 163} \\
Accuracy & & 0.79 & & & 0.85 & & \\
Macro Avg & 0.76 & 0.74 & 0.62 & 0.78 & 0.81 & 0.70 & \\
Weighted Avg & 0.88 & 0.79 & 0.77 & 0.91 & 0.85 & 0.85 & \\
\bottomrule
\end{tabular}
\caption{Detailed performance metrics by neighborhood for string matching and GPT-4 Mini classification methods. The Support column indicates the number of test samples for each neighborhood. Dashes indicate cases where precision could not be calculated due to zero predictions.}
\label{tab:detailed_classification_performance}
\end{table*}

\begin{table}[t]
\centering
\small
\begin{tabular}{lrrrr}
\toprule
\textbf{Variable} & \textbf{Coef.} & \textbf{Std. Err.} & \textbf{t} & \textbf{P>|t|} \\
\midrule
Intercept & 0.10 & 0.01 & 10.40 & 0.00 \\
Bedrooms & 0.02 & 0.00 & 8.83 & 0.00 \\
Bathrooms & 0.02 & 0.00 & 6.25 & 0.00 \\
Rent & 0.00 & 0.00 & -14.56 & 0.00 \\
Square Footage & 0.00 & 0.00 & 2.17 & 0.03 \\
Topic 2 Rental & 0.03 & 0.01 & 3.40 & 0.00 \\
Topic 3 Property  & 0.20 & 0.01 & 14.99 & 0.00 \\
Topic 4 Spanish & 0.10 & 0.01 & 9.42 & 0.00 \\
Topic 5 Amenities & -0.03 & 0.01 & -2.98 & 0.00 \\
Topic 6 Condition & 0.02 & 0.01 & 1.86 & 0.06 \\
Topic 7 Contact & 0.05 & 0.01 & 5.56 & 0.00 \\
\bottomrule
\end{tabular}
\caption{Regression results demonstrating how relative distance from the center of the neighborhood associates with rental unit characteristics.}
\label{tab:regression_results}
\end{table}

\subsection{Topic Modeling Results}

Working with more than 35,000 unique and often verbose advertisements, we fit a topic model on $k=25$ topics to identify main themes in the rental listings. We do not provide interpretations of the topic, but show the common words in Table \ref{tab:topic_words25}.

\begin{table}[t]
\setlength{\tabcolsep}{2pt}
\small
\centering
\begin{tabular*}{\columnwidth}{@{\extracolsep{\fill}}lp{0.8\columnwidth}@{}}
\multicolumn{2}{@{}c@{}}{\textbf{Table 2: Topics for $k=25$}} \\
\midrule
\textbf{Topic} & \textbf{Common Words} \\
\midrule
\textbf{Topic 1} & hyde, property, si, terrace, la, estos, con, mac, elli, street \\
\textbf{Topic 2} & large, space, floor, living, dining, closet, bedroom, heat, storage, central \\
\textbf{Topic 3} & contact, call, schedule, photo, tour, please, unit, info, show, actual \\
\textbf{Topic 4} & space, walk, home, lease, great, living, street, one, loft, large \\
\textbf{Topic 5} & real, estate, star, text, tour, community, view, lounge, apartment, finish \\
\textbf{Topic 6} & lease, mo, blueground, month, furnished, amenity, stay, offer, view, access \\
\textbf{Topic 7} & cat, feature, floor, call, dishwasher, one, fee, lakeview, n, dog \\
\textbf{Topic 8} & modern, heat, property, central, bus, appliance, call, gas, cat, air \\
\textbf{Topic 9} & apartment, rental, lake, property, place, hill, alquilar, cheap, new, grove \\
\textbf{Topic 10} & unit, special, availability, subject, dog, weight, onsite, please, price, various \\
\textbf{Topic 11} & view, center, lake, amenity, free, onsite, community, michigan, call, window \\
\textbf{Topic 12} & fee, credit, deposit, tenant, new, security, included, move, pay, pet \\
\textbf{Topic 13} & hyde, apartment, regent, horas, onsite, market, la, est, e, museum \\
\textbf{Topic 14} & apartment, rental, lake, property, place, hill, cheap, find, new, grove \\
\textbf{Topic 15} & studio, property, bjb, internet, complimentary, e, change, picture, subject, fitness \\
\textbf{Topic 16} & fee, mile, white, home, tile, neighborhood, make, company, new, tenant \\
\textbf{Topic 17} & housing, water, included, heat, opportunity, landstar, equal, group, act, feature \\
\textbf{Topic 18} & contact, show, info, price, availability, email, renovated, included, click, web \\
\textbf{Topic 19} & lake, bus, block, cta, walk, red, minute, away, stop, distance \\
\textbf{Topic 20} & info, contact, show, click, il, id, feature, friendly, text, n \\
\textbf{Topic 21} & amenity, view, center, pool, fitness, luxury, outdoor, lounge, garage, window \\
\textbf{Topic 22} & logan, banker, coldwell, square, blue, real, estate, please, opportunity, equal \\
\textbf{Topic 23} & hyde, village, height, center, drexel, grand, nuestros, river, maintenance, m \\
\textbf{Topic 24} & fee, per, cat, lease, dog, application, deposit, one, heat, gas \\
\textbf{Topic 25} & new, appliance, stainless, steel, floor, central, large, feature, granite, dishwasher \\
\midrule
\end{tabular*}
\caption{Topic clusters from LDA topic modeling on Chicago Craigslist rental listings, $k=25$.}
\label{tab:topic_words25}
\end{table}

\begin{figure}[t!]
    \centering
    \includegraphics[width=\linewidth]{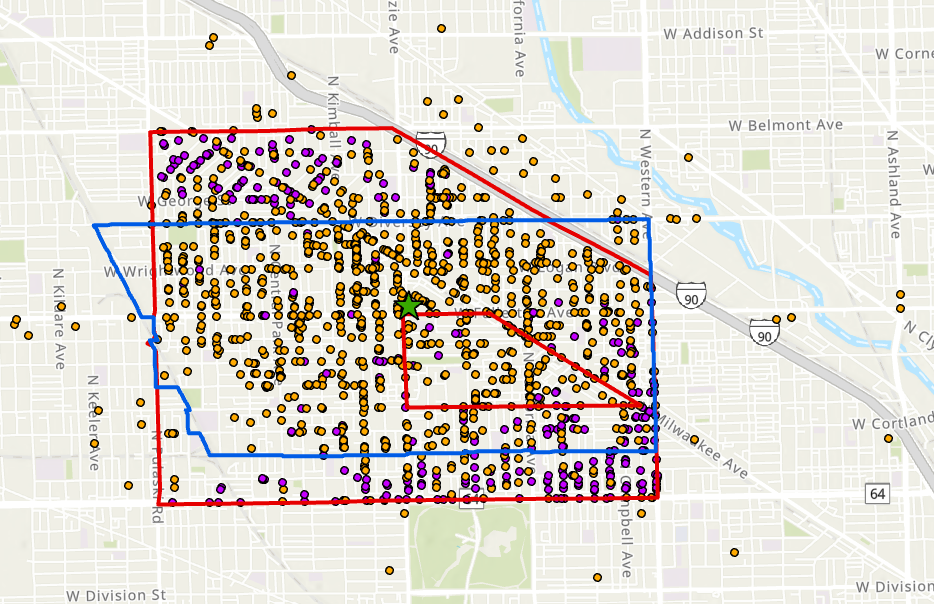}
    \caption{Contested boundaries for Logan Square neighborhood according to the official City of Chicago limits (blue), Zillow's definition (red), and claims in Craigslist rental listings (orange points for listings claiming to be in the neighborhood, purple if claiming to be elsewhere); the center of our Craigslist-defined neighborhood is represented by a green star. Logan Square represents a popular neighborhood with many postings and a number of nearby listings claiming to be within the neighborhood.}
    \label{fig:logan}
\end{figure}

\end{document}